\documentclass[conference]{IEEEtran}
\IEEEoverridecommandlockouts
\usepackage{cite}
\usepackage{amsmath,amssymb,amsfonts}
\usepackage{algorithmic}
\usepackage{graphicx}
\usepackage{textcomp}
\usepackage{xcolor}
\usepackage{algorithmic}
\usepackage{algorithm}
\usepackage{caption}
\usepackage{caption}
\def\BibTeX{{\rm B\kern-.05em{\sc i\kern-.025em b}\kern-.08em
    T\kern-.1667em\lower.7ex\hbox{E}\kern-.125emX}}

\begin{document}

\title{Learning to Control Dynamical Agents via Spiking Neural Networks and Metropolis-Hastings Sampling \\

\thanks{A. Safa supervised the project as Principal Investigator, contributed to the technical developments and to the writing of the manuscript. F. Mohsen contributed to the technical analysis and to the writing of the manuscript. A. Al-Zawqari contributed to the writing of the manuscript and provided technical feedback.}
}

\author{
\IEEEauthorblockN{Ali Safa$^1$, Farida Mohsen$^1$, Ali Al-Zawqari$^2$}
\IEEEauthorblockA{\textit{$^1$College of
Science and Engineering, Hamad Bin Khalifa University, Doha, Qatar}\\
\textit{$^2$ Department of Fundamental Electricity and Instrumentation, Vrije Universiteit Brussel, Brussels, Belgium}\\
asafa@hbku.edu.qa, fmohsen@hbku.edu.qa, aalzawqa@vub.be}

}

\maketitle

\begin{abstract}

Spiking Neural Networks (SNNs) offer biologically inspired, energy-efficient alternatives to traditional Deep Neural Networks (DNNs) for real-time control systems. However, their training presents several challenges, particularly for reinforcement learning (RL) tasks, due to the non-differentiable nature of spike-based communication. In this work, we introduce what is, to our knowledge, the first framework that employs Metropolis-Hastings (MH) sampling, a Bayesian inference technique, to train SNNs for dynamical agent control in RL environments without relying on gradient-based methods. Our approach iteratively proposes and probabilistically accepts network parameter updates based on accumulated reward signals, effectively circumventing the limitations of backpropagation while enabling direct optimization on neuromorphic platforms. We evaluated this framework on two standard control benchmarks: AcroBot and CartPole. The results demonstrate that our MH-based approach outperforms conventional Deep Q-Learning (DQL) baselines and prior SNN-based RL approaches in terms of maximizing the accumulated reward while minimizing network resources and training episodes. 
\end{abstract}



\section{Introduction}

Efficient control of dynamic agents is fundamental in various contemporary artificial intelligence (AI) applications, including robotics, autonomous driving, and industrial automation. Such agents typically leverage Reinforcement Learning (RL) methods, which iteratively learn optimal control strategies by interacting with their environments and maximizing accumulated rewards over time \cite{sutton2018reinforcement}. Recent advances in deep RL, particularly methods such as deep Q-Learning (DQL) and Proximal Policy Optimization (PPO), have achieved impressive performance on complex decision-making tasks \cite{mnih2015human, schulman2017ppo}. However, these approaches suffer from substantial drawbacks, including intensive computation requirements, high energy demands, and fundamental dependence on gradient calculation through backpropagation, limiting their practical deployment on ultra-energy-constrained platforms and specialized mixed-signal neuromorphic hardware \cite{davies2018loihi}.

Spiking Neural Networks (SNNs) have emerged as a viable alternative to traditional deep learning models due to their biological realism and inherent efficiency \cite{ghosh2009spiking,maass1997networks, naya2021spiking}. SNNs utilize discrete spikes or action potentials for neuron communication, closely mirroring biological neural activity \cite{charlotte, hairball}. This spike-based communication mechanism inherently offers substantial energy efficiency, effective temporal dynamics processing, and enhanced biological plausibility, positioning SNNs as highly suitable for low-power real-time tasks on neuromorphic platforms \cite{safa2024attention, alea2024skin}. Despite these attractive features, efficiently training SNNs for RL remains challenging \cite{tavanaei2019deep}, mainly because spike events are non-differentiable, thus making traditional gradient-based learning methods, such as backpropagation, ineffective \cite{zenke2018superspike}. Various surrogate gradient methods have attempted to overcome these challenges but remain computationally intensive and less robust when implemented on analog neuromorphic systems, which frequently experience process-voltage-temperature (PVT) variations \cite{neftci2019surrogate}. Therefore, gradient-free optimization approaches are necessary to accommodate neuromorphic hardware constraints and maintain effective training performance.
\begin{figure}[t]
\centering
    \includegraphics[scale = 0.255]{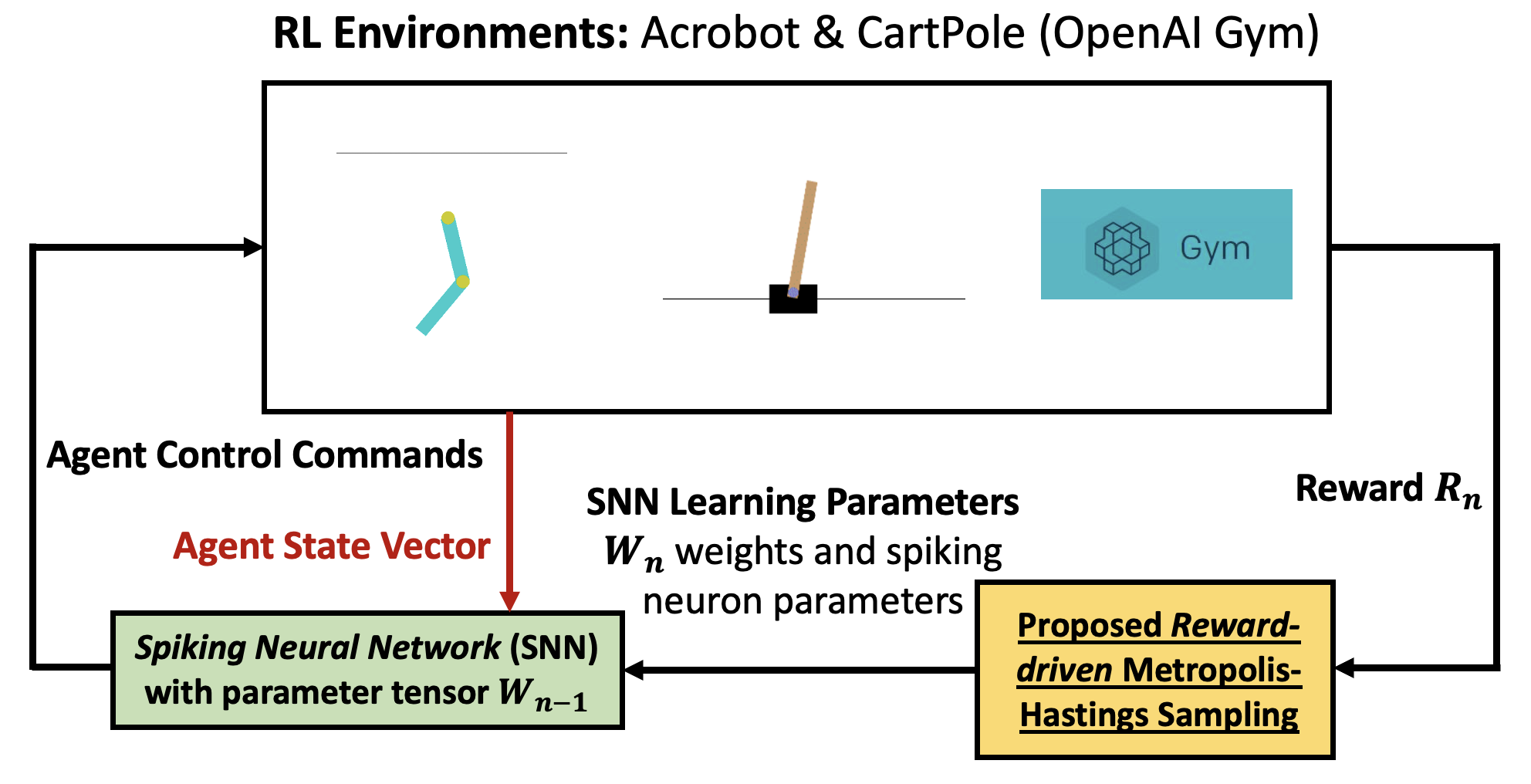}
    \caption{\textit{\textbf{Proposed Reward-driven Metropolis-Hastings SNN training approach.} A Spiking Neural Network (SNN) is used to infer agent control commands using the current state of the agent as input data. During the $n^{th}$ agent control episode, the accumulated reward $R_n$ is returned by the environment. The accumulated reward is then given to our proposed reward-driven Metropolis-Hastings procedure in order to infer a new SNN parameter tensor proposal $W_n$ (containing all the learnable SNN weights and neural parameters). The SNN parameters are then updated with the new set of parameters $W_n$ and the algorithm moves to its next iteration $n+1$.   }}
    \label{gagraphee}
\end{figure}

Motivated by recent developments in probabilistic neural computing, previous work have proposed the use of Metropolis-Hastings (MH) sampling \cite{chib1995understanding} as a gradient-free approach to train SNN signal \textit{classifiers} \cite{safa2024chiploop}. Indeed, MH sampling provides inherent robustness against hardware noise and analog circuit imperfections \cite{robert2010monte}. Unlike gradient-based methods, MH sampling optimizes neural network parameters by probabilistically accepting or rejecting updates based on their impact on a defined objective function, thereby directly addressing the non-differentiable nature of spike-based communication \cite{safa2024chiploop}. This capability enables chip-in-the-loop training, where SNNs are optimized directly on neuromorphic devices, accounting for PVT variations without the need for gradient information  \cite{charlotte, safa2024chiploop}. 

In this paper, we expand MH sampling methodology into the domain of \textit{RL for dynamical agent control}, presenting the first exploration of MH-trained SNN policies within RL contexts (see Fig. \ref{gagraphee}). The key contributions of this work are as follows:

\begin{enumerate} 
\item To our knowledge, this study is the first to employ MH sampling for training SNNs to control dynamic agents in RL scenarios. 
\item We demonstrate that our reward-driven MH-trained SNNs consistently surpass conventional DQL baselines on standard RL benchmarks, including AcroBot, and CartPole. 

\item Our MH-based SNN training method significantly enhances generalization capabilities compared to traditional gradient-based approaches used in DQL, achieving superior accuracy with simpler network architectures. 
\end{enumerate}

The remainder of the paper is structured as follows: Section \ref{background} outlines relevant background theory on MH sampling. Section \ref{methods} describes our proposed methodology for SNN-based dynamical agent control using MH sampling. Section \ref{experiments} presents comprehensive experimental evaluations. Finally, Section \ref{conclusion} summarizes our findings and outlines directions for future research.

\begin{figure*}[htbp]
\centering
    \includegraphics[scale = 0.48]{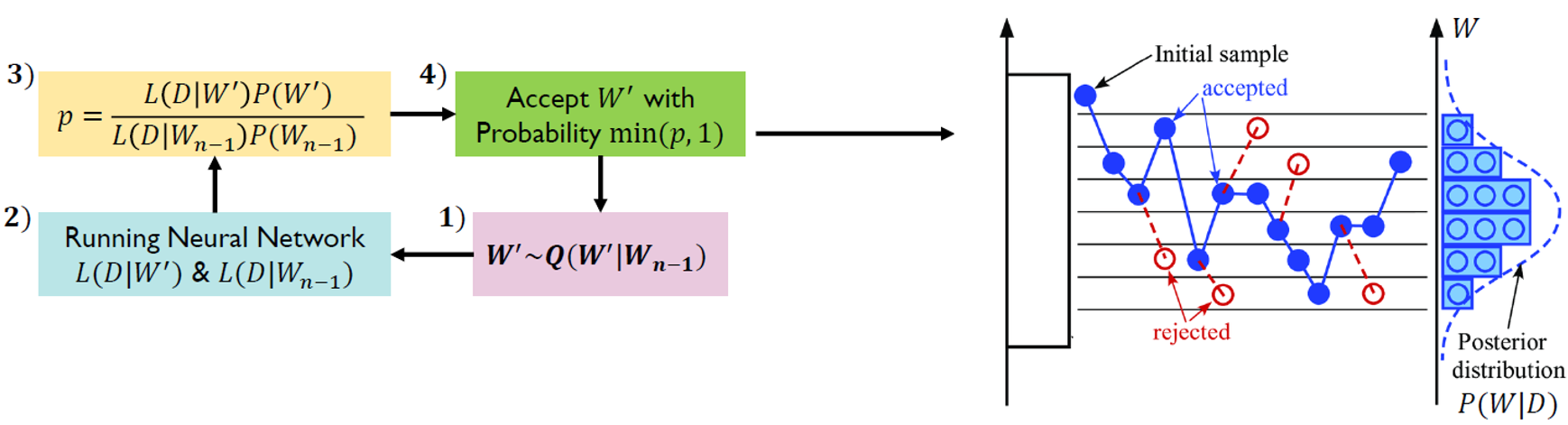}
    \caption{\textit{\textbf{Conceptual illustration of the Metropolis-Hastings algorithm.} }}
    \label{mhalg}
\end{figure*}

\section{Background Theory}
\label{background}
\subsection{Metropolis-Hastings sampling}
\label{mhsamplingbase}
In this work, we make use of the Metropolis-Hastings sampling algorithm \cite{robert2015mh} as a Bayesian training approach for learning the SNN weights. Metropolis-Hastings provides a computationally tractable way for estimating the posterior distribution of the SNN weights $P(W|D)$ given the data $D$ following Bayes' rule \cite{robert2015mh}:
\begin{equation}
    P(W|D) = \frac{L(D|W) P(W)}{P(D)}
    \label{bayes}
\end{equation}
where, in the context of this paper, $W$ denotes the SNN weights, $L(D|W)$ is the likelihood probability distribution of the data given the SNN weights, $P(W)$ is the prior distribution, $P(D)$ is the data distribution and $P(W|D)$ the posterior distribution of $W$. 

It is commonly known in the Bayesian inference literature that the \textit{explicit} use of the Bayes rule (\ref{bayes}) typically comes with a high computational complexity the more the dimensions of the weight tensor $W$ grows \cite{bayesiandnn}. For this reason, we make use of the Metropolis-Hastings sampling algorithm for \textit{implicitly} estimating the posterior density over the weights $P(W|D)$. Metropolis-Hastings estimates $P(W|D)$ by providing a sequence of weight tensor samples $W \sim P(W|D)$ which provably correspond to true samples drawn from the distribution $P(W|D)$ \cite{robert2015mh}. Remarkably, these samples are obtained without an exact knowledge of the posterior distribution $P(W|D)$. 

Fig. \ref{mhalg} provides a conceptual illustration of the Metropolis-Hastings sampling algorithm. First, the sampler is initialized with a random weight tensor $W_1$ at sampling step $n=1$. Then, a symmetric distribution $Q$ (typically a Gaussian) is used to obtain a proposal sample $W^{'} \sim Q(W^{'}|W_{n-1})$ using $W_{n-1}$ as the mean value of $Q$. Second, the likelihoods of the data $D$ given the newly-proposed weight $W^{'}$ and the previous weight $W_{n-1}$ are computed as $L(D|W^{'}), L(D|W_{n-1})$ (where $L$ is an arbitrary-chosen likelihood function suitable for the task to be solved). Third, the following ratio $p$ is computed:
\begin{equation}
    p = \frac{L(D|W^{'} P(W^{'})}{L(D|W_{n-1}) P(W_{n-1})}
    \label{ratiop}
\end{equation}
where $P(W^{'})$ and $P(W_{n-1})$ respectively denote the prior probability of $W^{'}$ and $W_{n-1}$. Finally, the newly-proposed sample $W^{'}$ is accepted as a true sample drawn from $P(W|D)$ if and only if the outcome of a binary Bernoulli trial with probability $\min(p,1)$ is successful. 

\subsection{Spiking Neural Networks}

This work makes use of Spiking Neural Networks (SNNs) for controlling various dynamical agents within an RL scenario. SNNs form a type of neural network architecture that closely mimics the way information is processed in the brain \cite{safa2024attention}. Compared to conventional Deep Neural Networks (DNNs), SNNs make use of neurons which fire binary spikes as their output (i.e., action potentials). As spiking neuron model, the experiments provided in this paper make use of \textit{Leaky Integrate-and-Fire} (LIF) neurons which operate as follows \cite{safa2024attention}. Upon receiving their input current at time-step $m$:
\begin{equation}
    J[m]=\Bar{w}^T \Bar{s}[m]
\end{equation}
where $\Bar{w}$ is the weight vector associated to the neuron, $T$ denotes the transpose and $\Bar{s}$ is the set of input spikes coming from the previous layer, LIF neurons integrate the current $J[m]]$ into their membrane potential $V$ as:
\begin{equation}
    V[m] \xleftarrow{} \alpha V[m-1] + (1-\alpha) J[m]
\end{equation}
where $\alpha$ is the membrane leakage constant. As soon as $V[m]$ crosses a threshold $\mu$, the neuron emits an output spike $\delta_{out} = 1$ and the membrane potential $V[m] \xleftarrow{} 0$ is reset to zero.

Using the LIF neuron model as non-linearity instead of classical analog neurons such as ReLU, it is possible to build neural network architectures that communicate via action potentials or spikes and that mimic in a more faithful manner the neural dynamics taking place in biological brains \cite{charlotte}.

In the next Section, MH sampling (covered in Section \ref{mhsamplingbase}) will be leveraged to build a \textit{reward-driven} SNN training method for learning the control of dynamical agents within various Reinforcement Learning (RL) scenarios.

\section{Methods}
\label{methods}

In this paper, we propose to leverage MH sampling for learning to control dynamical agents within a RL-like setting. In such RL setting, the environment in which the dynamical agent operates returns rewards $r_j$ for each time step $j$ during the execution of an agent control episode, with a maximum number of time step per episode $j=1,...,T_{max}$ during which the agent is expected to attain its goal. It is assumed that the agent control is done through an arbitrary SNN architecture and we note by $W$ the tensor containing all of the learnable SNN parameters (e.g., weights, neural membrane decay constant $\alpha_{decay}$ and spiking threshold $\mu$). Hence, the optimization objective of the SNN learning process can be written as the maximization of the accumulated reward:
\begin{equation}
    W_{best} = \arg \max_{W} \sum_{j=1}^{T_{max}} r_j(W)
    \label{learnobj}
\end{equation}

Here, we note a similarity between the SNN learning objective (\ref{learnobj}) and the objective of MH sampling, which is to provide SNN parameter samples that will maximize a certain likelihood $L(D|W)$ in (\ref{bayes}). Hence, by defining:
\begin{equation}
    L \equiv R(W) = \sum_{j=1}^{T_{max}} r_j(W)
    \label{pseudolik}
\end{equation}
where $R(W)$ denotes the total accumulated reward over an episode, it is possible to use the pseudo-likelihood measure $L$ in (\ref{pseudolik}) in place of an exact likelihood function within the MH sampling procedure. This is possible thanks to the fact that MH sampling makes use of the \textit{ratio} between the current and past likelihood functions in (\ref{ratiop}), which makes it possible to drop the requirement for an exact knowledge of the normalization terms defining the \textit{absolute} likelihood function.  

Following these observations, our proposed reward-driven MH algorithm for SNN training is provided in Algorithm \ref{alg:metropolis_hastings}. The algorithm follows the MH sampling procedure illustrated in Fig. \ref{mhalg} with the crucial addition of the accumulate rewards $R(W)$ and $R(W_{n-1})$ as likelihood values in lines 4-5-6 of Algorithm \ref{alg:metropolis_hastings}.
\begin{algorithm}[h]
\caption{Proposed Reward-driven Metropolis-Hastings method for SNN training}
\label{alg:metropolis_hastings}
\begin{algorithmic}[1]
\REQUIRE Gym environment (e.g., \texttt{AcroBot} and \texttt{Cart-Pole}), initial weights $W_0$
\STATE Init. $R_{best} = -\infty$, $W_{best} = W_0$ 
\FOR{$n = 1$ to $N_{iter}$}
    \STATE Sample proposal $W' \sim Q(W'|W_{n-1})$
    \STATE Run the gym agent control environment with the SNN with weights $W'$ and compute the total accumulated reward $R(W')$.
    \STATE Run the gym agent control environment with the SNN with previous weights $W_{n-1}$ and compute the total accumulated reward $R(W_{n-1})$.
    \STATE Assign likelihoods: $L_1 \gets R(W')$, $L_2 \gets R(W_{n-1})$
    \STATE Compute priors: $P_1 \gets P(W')$, $P_2 \gets P(W_{n-1})$
    \STATE Compute acceptance ratio: $p = \frac{L_1 \cdot P_1}{L_2 \cdot P_2}$

    \STATE Sample $u \sim \text{Uniform}(0, 1)$
    \IF{$u < \min(p, 1)$}
        \STATE Accept proposal: $W_n \gets W'$
    \ELSE
        \STATE Reject proposal: $W_n \gets W_{n-1}$
    \ENDIF

    \IF{$R(W') > R_{best} $}
    \STATE $R_{best} \gets R(W')$
    \STATE $W_{best} \gets W_n$
    \ENDIF
    
\ENDFOR
\STATE Return $W_{best}$
\end{algorithmic}
\end{algorithm}

In the next section, we apply the reward-driven MH-based SNN training method (Algorithm \ref{alg:metropolis_hastings}) to several popular agent control environments. Then, we compare its performance with Deep Q-learning implemented using conventional (non-spiking) DNNs with ReLU activation functions.
\begin{figure}[htbp]
\centering
    \includegraphics[scale = 0.8]{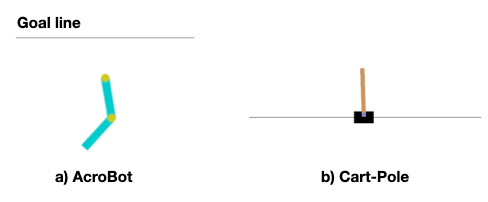}
    \caption{\textit{\textbf{Environments used in our experiments for the control of dynamical agents.} a) \texttt{AcroBot}, b) \texttt{Cart-Pole} }}
    \label{envs}
\end{figure}

\section{Experiments}
\label{experiments}

In this Section, we assess our proposed method on two different \textit{continuous control} environments of the \textit{Open AI Gym Suite} \cite{brockman2016openai}: \texttt{AcroBot}, and \texttt{Cart-Pole} (see Fig. \ref{envs}). Throughout the experiments, we compare the proposed SNN-MH control system with \textit{Deep Q-Learning} (DQL), as a typical reinforcement learning technique used to solve control environments in literature \cite{mnih2015human}. 

We base our DQL setups on the openly available implementation provided in the link below\footnote{\texttt{https://github.com/johnnycode8/gym\_solutions}}. For DQL training, we consistently use the Adam optimizer with its default hyper-parameters. Standard experience replay and $\epsilon$-greedy policy sampling is used during each DQL training as well \cite{Plaat_2022}. As for the other DQL tuning hyper-parameters, we found that the base parameters provided in the implementation below$^1$ were already satisfying in terms of convergence speed and accumulated reward, since changing the hyper-parameters further did not improve the results in our case.  

\begin{figure}[htbp]
\centering
    \includegraphics[scale = 0.28]{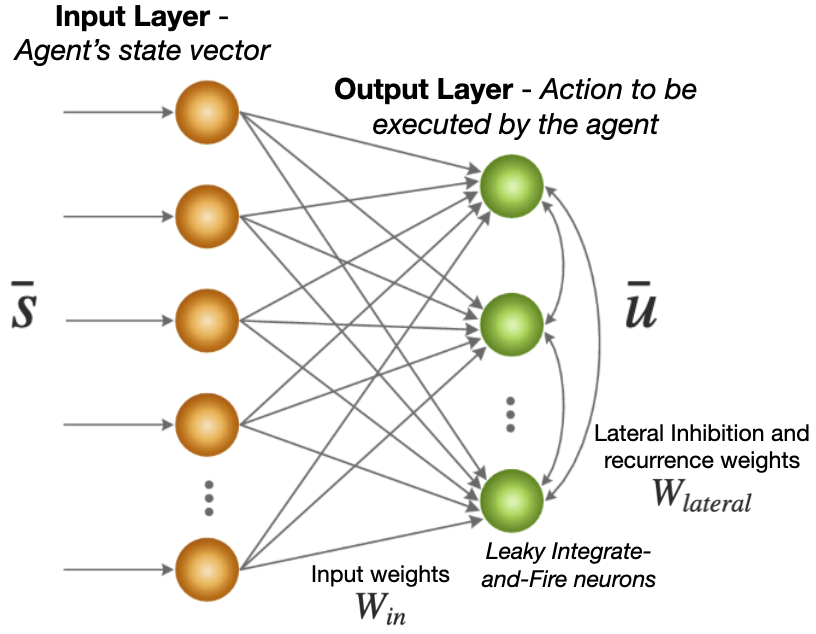}
    \caption{\textit{\textbf{SNN architecture.} We use a 1-layer SNN topology with input weight matrix $W_{in}$ and lateral recurrence weight matrix $W_{lateral}$. As input to the network, we feed the agent's state vector (with the input dimension depending on the environment to be solved). The output is then used to select the action to be executed by the agent (each output neuron corresponding to a possible action).}}
    \label{snntopo}
\end{figure}

As SNN architecture, we use the 1-layer topology shown in Fig. \ref{snntopo}. The SNN has two sets of weights: an input weight matrix $W_{in}$ mapping the agent's state vector $\Bar{s}$ to the output LIF neurons, and a lateral inhibition and recurrence weight matrix $W_{lateral}$ feeding back the output spike vector $\Bar{u}$ to the output neurons. The decay constant $\alpha_{decay}$ and threshold $\mu$ of the output LIF neurons are also tuned by the proposed MH training procedure. Naturally, the input and output dimensions depend on the specifications of each environment to be solved (\texttt{AcroBot} having $6$ states, and \texttt{Cart-Pole} having $4$ states.

During the training procedure, the MH algorithm provides samples of vectors containing an estimate of the network parameters $\{W_{in}, W_{lateral}, \alpha_{decay}, \mu\}$ with the goal of maximizing the accumulated reward returned by the environment (see Algorithm \ref{alg:metropolis_hastings}).
\setcounter{figure}{4}
\subsection{AcroBot Environment}

The \texttt{AcroBot} environment consists of a double pendulum that needs to be controlled in order for its tip to reach a goal line placed above the pendulum (see Fig. \ref{envs} a). The agent has $6$ observable states related to the angular position of the pendulum joints and their angular velocities. 

\begin{figure}[htbp]
\centering
    \includegraphics[scale = 0.52]{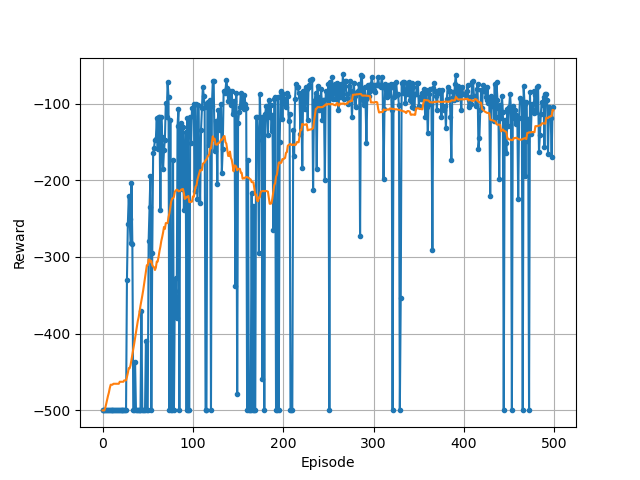} 
    \caption{\textit{\textbf{Evolution of the accumulated reward along the MH sampling episodes.} Using the \texttt{AcroBot} environment. The orange line was obtained by applying a moving average filter of width $50$ on the reward curve in blue.}}
    \label{acrobot_reward}
\end{figure}

Fig. \ref{acrobot_reward} shows the evolution of the accumulated reward during the MH training procedure on the \texttt{AcroBot} environment. It can be seen that after an initial set of around 160 episodes, the accumulated reward grows before reaching a plateau around $-100$ (denoting that the agent can successfully solve the environment within around $100$ episodes    ).

\begin{figure}[htbp]
\centering
    \includegraphics[scale = 0.54]{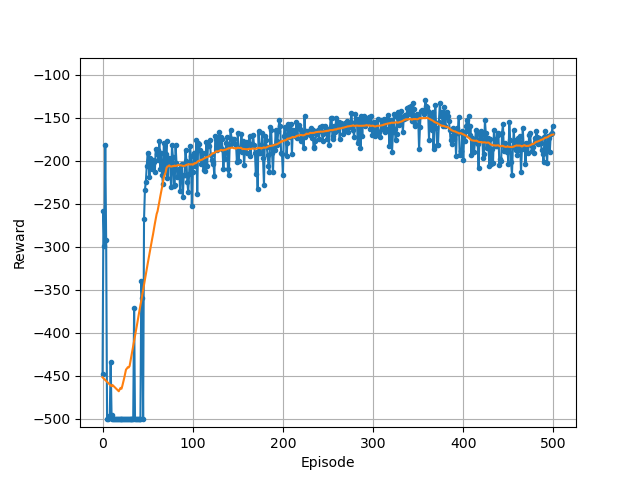}
    \caption{\textit{\textbf{Deep Q-Learning, evolution of the accumulated reward.} Using the \texttt{AcroBot} environment. The orange line was obtained by applying a moving average filter of width $50$ on the reward curve in blue. Compared to our proposed SNN-MH setup, the DQL setup achieves a significantly smaller accumulated reward of $\sim -150$ vs. $\sim -100$ in the SNN-MH case (see Fig. \ref{acrobot_reward}).  }}
    \label{acrobot_DQL}
\end{figure}

In addition, Fig. \ref{acrobot_DQL} shows the results obtained using the DQL baseline (using a similar DNN topology to the SNN architecture shown in Fig. \ref{snntopo}, with the LIF neurons replaced with ReLUs). It can be seen that the DQL setup achieves a reward plateau with a significantly smaller accumulated reward of at most $\sim -150$, while our proposed SNN-MH setup achieves an accumulated reward of $\sim -100$ (see Table \ref{tab:row_col_titles}). 

\subsection{Cart-Pole Environment}
The \texttt{Cart-Pole} environment consists of a cart that can move horizontally with an inverted pendulum mounted on its top (see Fig. \ref{envs} b). The goal is to control the cart in order to keep the inverted pendulum vertical and not let it fall due to gravity. The agent has $4$ observable states related to the position and velocity of the cart, as well as the angle and angular velocity of the inverted pendulum. 

\begin{figure}[htbp]
\centering
    \includegraphics[scale = 0.53]{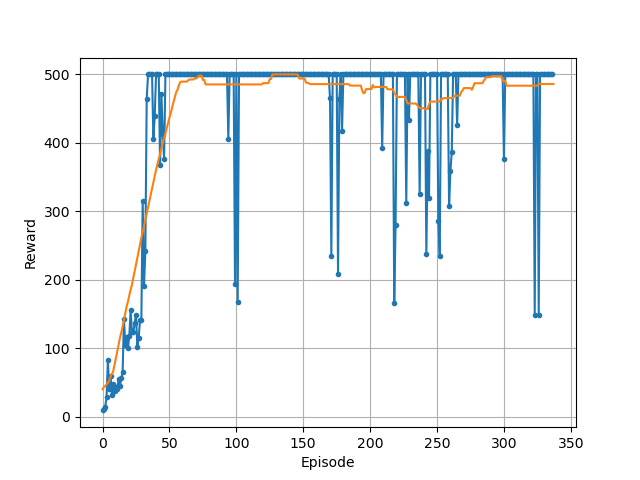}
    \caption{\textit{\textbf{Evolution of the accumulated reward along the MH sampling episodes.} Using the \texttt{Cart-Pole} environment. The orange line was obtained by applying a moving average filter of width $50$ on the reward curve in blue.}}
    \label{cartpole_reward}
\end{figure}

Fig. \ref{cartpole_reward} shows the evolution of the accumulated reward during the MH training procedure on the \texttt{Cart-Pole} environment. It can be seen that the accumulated reward grows and reaches within $50$ episodes a reward plateau around $500$ (which is the maximum reward attainable in the case of the \texttt{Cart-Pole} environment).

\begin{figure}[htbp]
\centering
    \includegraphics[scale = 0.56]{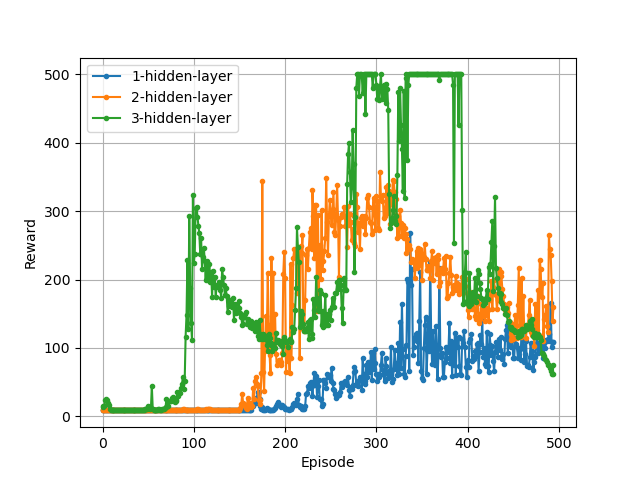}
    \caption{\textit{\textbf{Deep Q-Learning, evolution of the accumulated reward.} Using the \texttt{Cart-Pole} environment. In contrast to our proposed SNN-MH setup, using a 1- or even 2-hidden layer network in the DQL case is not enough for solving the environment (i.e, reaching the maximum reward of $500$).}}
    \label{cartpole_DQL}
\end{figure}

In addition, Fig. \ref{cartpole_DQL} shows the results obtained using the DQL baseline. Remarkably, DQL was not able to solve the \texttt{Cart-Pole} environment using a similar DNN topology as used by our SNN (i.e., a 1-layer network). This is why we experimented with an increasing amount of hidden layers in the DNN (using $16$ neurons in each hidden layer). Fig. \ref{cartpole_DQL} shows that the maximum accumulated reward grows as we increase the number of hidden layers. At the 3 hidden layers, the DQL setup is finally able to solve the \texttt{Cart-Pole} environment by reaching an accumulated reward of $500$. This is in striking contrast with our proposed SNN-MH setup which is able to solve \texttt{Cart-Pole} by reaching a reward plateau of $500$ (see Fig. \ref{cartpole_reward}) while using the 1-layer SNN architecture described in Fig. \ref{snntopo}. The comparison between the SNN-MH setup and the DQL setup is also reported in Table \ref{tab:row_col_titles}.

\linespread{1}

\begin{table}[h]
\centering
\begin{tabular}{|l|l|l|}
\hline
 & \textbf{SNN-MH} & \textbf{DQL} \\
\hline
\texttt{AcroBot} & $\mathbf{-90}$ & $-140$ \\
\texttt{Cart-Pole} & $\mathbf{500}$ & $280$ \\
\hline
\end{tabular}
\linespread{0.82} 
\caption{\textit{\textbf{Maximum accumulated reward} for both the SNN-MH and the DQL with equivalent 1-hidden-layer architecture (rewards are rounded to the nearest decade).}}
\label{tab:row_col_titles}
\end{table}
\linespread{0.82} 

\subsection{Comparative Analysis with Prior SNN-based Methods}

To contextualize our MH-trained SNN results beyond DQL baselines, we compare their performance on the CartPole and Acrobot environments with several representative SNN-based RL methods, as summarized in Table~\ref{tab:carot} and Table~\ref{tab:acrobot}.  The comparison is made in terms of \textit{i)} the number of neurons (the lower the better); \textit{ii)} the training efficiency (i.e., the number of episodes needed for convergence, the lower the better); and \textit{iii)} the accumulated reward value (the higher the better). 

For the Acrobot environment (see Table~\ref{tab:acrobot}), our MH-SNN demonstrates a drastic improvement in resource efficiency and learning speed. By using only $9$ neurons, which is over 95\% lower than the $204$ neurons used by Tihomirov et al. \cite{tihomirov2024actor}, our network converges roughly 20\% faster (160 vs.\ 200 episodes) and attains a 47\% higher best-case reward (–90 vs.\ –170). Similarly, compared to the DSQN approach \cite{akl2021porting}, which employs 521 neurons and requires 720 episodes to reach an average plateau near –100, our method delivers comparable or better performance in under a quarter of the training time and with more than 98\% fewer neurons. This clearly demonstrates the strength of our approach over these prior works. 

In the CartPole environment (see Table~\ref{tab:carot}), our MH-SNN achieves superior performance with a significantly lower number of neurons (6 neurons) compared to the previously proposed methods ~\cite{chung2020feedbacktdstdp, akl2021porting, markov}. Our approach converges within 50 episodes regarding learning efficiency, placing it among the fastest learners after the method of Markov et al.~\cite{markov} which uses a considerably larger network of 224 neurons compared to our work. Finally, our model reaches the maximum reward of 500 on the CartPole-v1 environment.  While the Feedback-modulated TD-STDP method proposed in~\cite{chung2020feedbacktdstdp} also reports a similar reward, it requires a substantially larger architecture and a larger number of training episodes. In addition, DSQN ~\cite{akl2021porting} and the model proposed by Markov et al.~\cite{markov} report a plateau value of $\sim$200 after convergence, due to the limited number of $200$ time steps for CartPole-v0 (vs. 500 times steps for CartPole-v1 as used in this work). 
Even though not directly comparable due to their use of a \textit{customized} CartPole environment, Plank et al.~\cite{plank2025cartpolebenchmark} report up to 14,985 successful agent control time steps over a total simulation time of 15,000 time steps (leading to an accumulated reward of 14,985). Even though their long-horizon CartPole task is not directly comparable to the standard episodic setup used in most prior works and in our work, their proposed method falls short in terms of episodes needed for convergence, since they report the need for $40$ training epochs with each epoch composed of 15,000 time steps, leading to a total number of $40 \times 15,000 = 600,000$ training time steps needed for convergence (vs. $50 \times 500 = 25,000$ in our work). In sum, our proposed method outperforms prior approaches in terms of number of neurons while achieving the maximum attainable reward of $500$ for CartPole-v1, and while being among the top performers in terms of convergence speed.

\begin{table*}[htbp]
    \centering
    \caption{\textit{\textbf{Acrobot Environment.} Proposed MH-SNN method vs. related SNN-based RL methods.}}
    \label{tab:acrobot}
    \begin{tabular}{|l|c|c|c|}
        \hline
        \textbf{Method} & \textbf{Number of neurons} & \textbf{Episodes to convergence} & \textbf{Accumulated reward} \\
        \hline
        
         Our MH-SNN  & \textbf{9 neurons}  &  \textbf{160} & \textbf{-90} \\
        \hline
       Tihomirov et al. ~\cite{tihomirov2024actor} & 204 & 200 & -170 
       \\
        
        \hline
       DSQNs  ~\cite{akl2021porting} & 521 neurons (6 input, 256 hidden, 256 hidden, 3 output) & 720 & -95 to -105 
       \\
        \hline

    \end{tabular}
\end{table*}

\begin{table*}[htbp]
    \centering
    \caption{\textit{\textbf{CartPole Environment.} Proposed MH-SNN method vs. related SNN-based RL approaches.}}
    \label{tab:carot}
        \begin{tabular}{|p{4.5cm}|p{6cm}|p{2cm}|p{4cm}|}
        \hline

        \textbf{Method} & \textbf{Number of neurons} & \textbf{Episodes to convergence} & \textbf{Accumulated reward} \\
        \hline
        Our MH-SNN  & \textbf{6 neurons} (4 input, 2 output) & 50  & \textbf{500} (CartPole-v1, max attainable) \\ \hline
        Feedback-modulated TD-STDP~\cite{chung2020feedbacktdstdp} & 60 neurons (40 critic, 20 actor) & 169.5 & \textbf{500} (CartPole-v1, max attainable) \\
        \hline
       Deep Spiking Q-Networks (DSQNs)  ~\cite{akl2021porting} & 134 neurons (4 input, 64 hidden, 64 hidden, 2 output) & 620 & \textbf{200} (CartPole-v0, max attainable) \\
        \hline
        Markov et al. \cite{markov} & 224 neurons & \textbf{10}
  & \textbf{200} (CartPole-v0, max attainable) \\
        \hline
         Plank et al. ~\cite{plank2025cartpolebenchmark} &  $\leq 12$ neurons ( networks range from 7 to 12 ) & 40 training epochs & \underline{14,985} (\textbf{custom} CartPole  environment with 15,000 time steps) \\ \hline

        \hline
    \end{tabular}
\end{table*}

\subsection{Discussion on the results}


Our findings show the effectiveness of training SNNs with MH sampling in controlling dynamic agents within RL environments. Firstly, MH sampling improves generalization compared to the gradient-based optimization used in DQL by exploring the Bayesian posterior and thus better avoiding local optima \cite{robert2015mh,jospin2022bayesian}. Concurrently, the event-driven, sparse activations of SNNs act as an inductive bias, regularizing the policy network and simplifying decision boundaries. This synergy underlies our method’s fast convergence and high reward performance on both AcroBot and CartPole. Table \ref{tab:row_col_titles} clearly demonstrates the usefulness of
our SNN-MH setup, where the reward-driven MH sampling method proposed in this work systematically reaches a higher accumulated reward than DQL.

Importantly, MH sampling enables effective policies with minimal network complexity. The Cart-Pole experiment shows that our single-layer SNN with just six neurons converges in approximately 50 episodes to the maximum reward of 500, outperforming the DQL approach, which required at least three layers to attain similar success (see Fig. 8). In addition, our method also outperforms prior SNN-based RL approaches in terms of network size and convergence speed (see Table~\ref{tab:acrobot} and Table~\ref{tab:carot}). Remarkably, our proposed method requires over $95\%$ less neurons and over $20\%$ lower convergence time in order to solve the Acrobot problem with a higher reported reward, compared to prior SNN-based RL works (see Table~\ref{tab:acrobot}). 

Finally, the \textit{gradient-free} nature of the MH sampling method used in this work confers robustness to hardware non-idealities during chip-in-the-loop training \cite{safa2024chiploop}. Its MH stochastic acceptance mechanism tolerates analog noise, voltage fluctuations, and timing jitter, making it particularly suitable for neuromorphic chip-in-the-loop implementations. 

\color{black}



%

\section{Conclusion}
\label{conclusion}

This paper has presented a novel reward-driven gradient-free approach for training SNNs using MH sampling in reinforcement learning tasks. Our framework successfully trains SNNs to control dynamical agents across three benchmark environments without backpropagation, making it suitable for training emerging types of analog and mixed-signal neuromorphic hardware. Experimental results demonstrate the proposed MH-based approach achieves either superior or on-par performance compared to DQL baselines and prior SNN-based RL methods. Future work will focus on developing adaptive sampling strategies as well as exploring the use of Hamiltonian Monte Carlo methods in order to scale the proposed reward-driven MH algorithm to more complex environments.

\bibliographystyle{unsrt}       
\bibliography{references}       

\end{document}